\documentclass[10pt,twocolumn,letterpaper]{article}

\usepackage{iccv}
\usepackage{times}
\usepackage{graphicx}
\usepackage{amsmath}
\usepackage{amssymb}

\usepackage{lipsum}

\usepackage{multicol}
\usepackage{multirow}
\usepackage{makecell}
\usepackage{booktabs}
\usepackage{color}
\usepackage{float}
\usepackage[accsupp]{axessibility}  

\usepackage[font=small]{caption}
\usepackage[font=small]{subcaption}

\makeatletter
\renewcommand\normalsize{%
\@setfontsize\normalsize\@xpt\@xiipt
\abovedisplayskip 2\p@ \@plus2\p@ \@minus5\p@
\abovedisplayshortskip \z@ \@plus3\p@
\belowdisplayshortskip 6\p@ \@plus3\p@ \@minus3\p@
\belowdisplayskip \abovedisplayskip
\let\@listi\@listI}
\makeatother

\usepackage[pagebackref=true,breaklinks=true,letterpaper=true,colorlinks,bookmarks=false]{hyperref}

\iccvfinalcopy 

\def\iccvPaperID{5660} 

\ificcvfinal\fi

\begin{document}

\title{End-to-End Dense Video Captioning with Parallel Decoding}


\author{
Teng Wang$^{1,2}$, Ruimao Zhang$^{3,4}$, Zhichao Lu$^2$, Feng Zheng$^{2*}$, Ran Cheng$^2$, Ping Luo$^1$ \\
$^1$ The University of Hong Kong \ $^2$ Southern University of Science and Technology  \\ $^3$ The Chinese University of Hong Kong (Shenzhen)\  $^4$ Shenzhen Research Institute of Big Data \\
{\tt\small tengwang@connect.hku.hk\ \  ruimao.zhang@ieee.org\ \  luzhichaocn@gmail.com} \ \ \\ {\tt\small f.zheng@ieee.org\ \ ranchengcn@gmail.com\ \ pluo@cs.hku.hk}
}


\maketitle

\ificcvfinal\thispagestyle{empty}\fi

\definecolor{urlcolor}{RGB}{231,72,151}

\begin{abstract}

Dense video captioning aims to generate multiple associated captions with their temporal locations from the video. Previous methods follow a sophisticated ``localize-then-describe'' scheme, which heavily relies on numerous hand-crafted components. 
In this paper, we proposed a simple yet effective framework for end-to-end dense video captioning with parallel decoding (PDVC), by formulating the dense caption generation as a set prediction task.
In practice, through stacking a newly proposed event counter on the top of a transformer decoder, the PDVC precisely segments the video into a number of event pieces under the holistic understanding of the video content, which effectively increases the coherence and readability of predicted captions.
Compared with prior arts, the PDVC has several appealing advantages:
(1) Without relying on heuristic non-maximum suppression or a recurrent event sequence selection network to remove redundancy, PDVC directly produces an event set with an appropriate size;
(2) In contrast to adopting the two-stage scheme, we feed the enhanced representations of event queries into the localization head and caption head in parallel, making these two sub-tasks deeply interrelated and mutually promoted through the optimization;
(3) Without bells and whistles, extensive experiments on ActivityNet Captions and YouCook2 show that PDVC is capable of producing high-quality captioning results, surpassing the state-of-the-art two-stage methods when its localization accuracy is on par with them. Code is available at 
{\url{https://github.com/ttengwang/PDVC}.}

\let\thefootnote\relax\footnotetext{* Corresponding author}

\end{abstract}

\section{Introduction}

\begin{figure}
\centering
\includegraphics[width=0.45\textwidth]{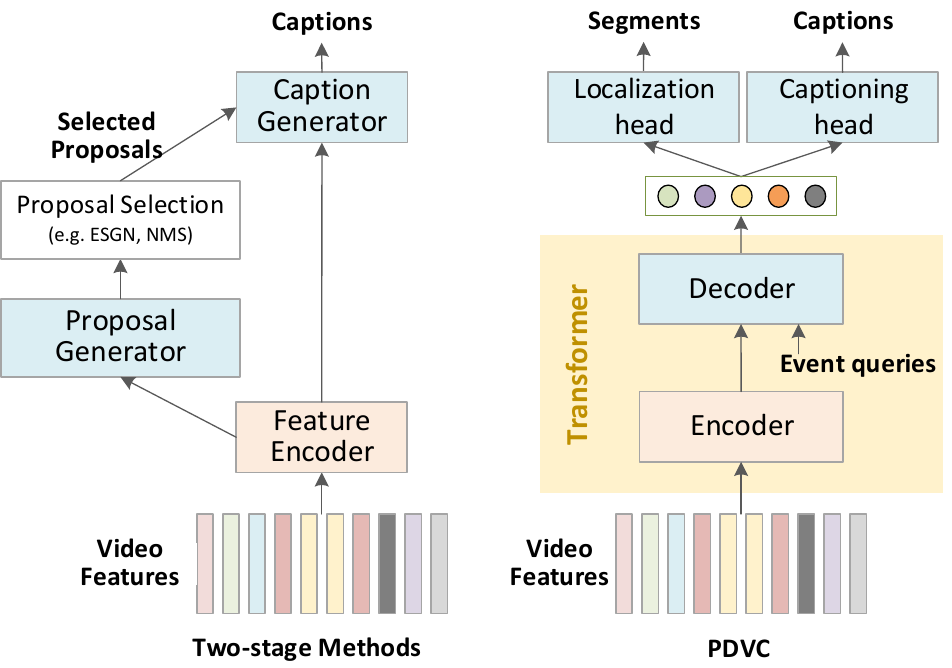}
\caption{The defacto two-stage pipeline vs. the proposed PDVC. The two-stage ``localize-then-describe" pipeline requires a dense-to-sparse proposal generation and selection process before captioning, which contains hand-crafted components and can not effectively exploit the potential mutual benefits between localization and captioning. PDVC adopts the vision transformer to learn attentive interaction of different frames, where the learnable event queries are embedded to capture the relevance between the frames and the events. Two prediction heads run in parallel over query features, leveraging the mutual benefits between two tasks and improving their performance together.}
\label{fig:intro}
\end{figure}

As an emerging branch of video understanding, video captioning has received an increasing attention in the recent past ~\cite{gao2017video,wei2019Memory,qi2019sports,wang2018Reconstruction,chen2017video, pan2016jointly, rohrbach2013translating, venugopalan2014translating, venugopalan2015sequence}, aiming to generate a natural sentence to describe one main event of a short video.
However, since realistic videos are usually long, untrimmed, and composed of a variety of events with irrelevant background contents, 
the above single-sentence captioning methods tend to generate sentences of blandness with less information. 
To circumvent the above dilemma, dense video captioning (DVC)~\cite{krishna2017dense, li2018jointly, xu2019joint, DaS, Suin2020efficient} is developed for automatically localizing and captioning multiple events in the video, 
which could reveal detailed visual contents and generate the coherent and complete descriptions.

Intuitively, dense video captioning can be divided into two subtasks, termed event localization and event captioning. 
As shown in Fig.~\ref{fig:intro}, the previous methods usually solve this problem by a two-stage ``localize-then-describe" pipeline. 
It firstly predicts a set of event proposals with accurate boundaries.
By extracting fine-grained semantic cues and visual contexts of the proposal, the detailed sentence description is finally decoded by the caption generator. 
%
%
The above scheme is straightforward but suffers from the following issues:
1) By considering the captioning as the downstream task, the performance of such a scheme highly relies on the quality of the generated event proposals, which limits the mutual promotion of these two sub-tasks. 
2) The performance of proposal generators in previous methods depends on careful anchor design~\cite{krishna2017dense, zhou2018end, li2018jointly, wang2018bidirectional, Mun2019stream, wang2020event} and proposal selection post-processing (\eg, non-maximum suppression~\cite{krishna2017dense,  zhou2018end, li2018jointly, wang2018bidirectional, Mun2019stream, wang2020event}). 
These hand-crafted components introduce additional hyper-parameters that highly rely on manual thresholding strategies, hindering the progress toward a fully end-to-end captioning generation.

To tackle the above issues, this paper proposes a pure end-to-end dense Video Captioning framework with Parallel Decoding termed PDVC.
As shown in Fig.~\ref{fig:intro}, instead of invoking the two-stage scheme, we directly feed the intermediate representation used for proposal generation into a captioning head that is parallel to the localization head.
By doing so, PDVC aims to directly exploit inter-task association at the feature level.
The intermediate feature vectors and the target events could be matched in a one-to-one correspondence, making the feature representations more discriminative to identify a specific event.

In practice, we consider the dense video captioning task as a set prediction problem.
The proposed PDVC directly decodes the frame features, which are extracted from a Vision Transformer, into an event set with their locations and corresponding captions by applying two parallel prediction heads, \ie, localization head and captioning head. 
Since the appropriate size of the event set is an essential indicator for dense captioning quality~\cite{Mun2019stream, fujitasoda}, 
a newly proposed event counter is also stacked on the top of the Transformer decoder to further predict the number of final events.
By introducing such a simple module, PDVC could precisely segment the video into a number of event pieces under the holistic understanding of the video content,
avoiding the information missing as well as the replicated caption generation caused by unreliable event number estimation.

We evaluate our model on two large-scale video benchmarks, ActivityNet Captions and YouCook2. Even with a lightweight caption head (vanilla LSTM), our method can achieve comparable performance against state-of-the-art methods which adopts well-designed attention-based LSTM~\cite{wang2018bidirectional, wang2020event} or Transformer~\cite{zhou2018end}. In addition, we show quantitatively and qualitatively that the generated proposals gain benefit from the paralleling decoding design. Even with a weakly supervised setting (without location annotations), we show our model can implicitly learn the location-aware features from captions.

To summarize, the major contributions of this paper are three folds.
1) We propose a novel end-to-end dense video captioning framework named PDVC by formulating DVC as a parallel set prediction task, significantly simplifying the traditional pipeline which highly depends on hand-crafted components.
2) We further improve PDVC with a novel event counter to estimate the number of events in the video, greatly increasing the readability of generated captions by avoiding the unrealistic event number estimation.
3) Extensive experiments on ActivityNet Captions and YouCook2 show state-of-the-art performance over existing methods.

\section{Related Work}

\begin{figure*}
    \centering
    \includegraphics[width=1. \textwidth]{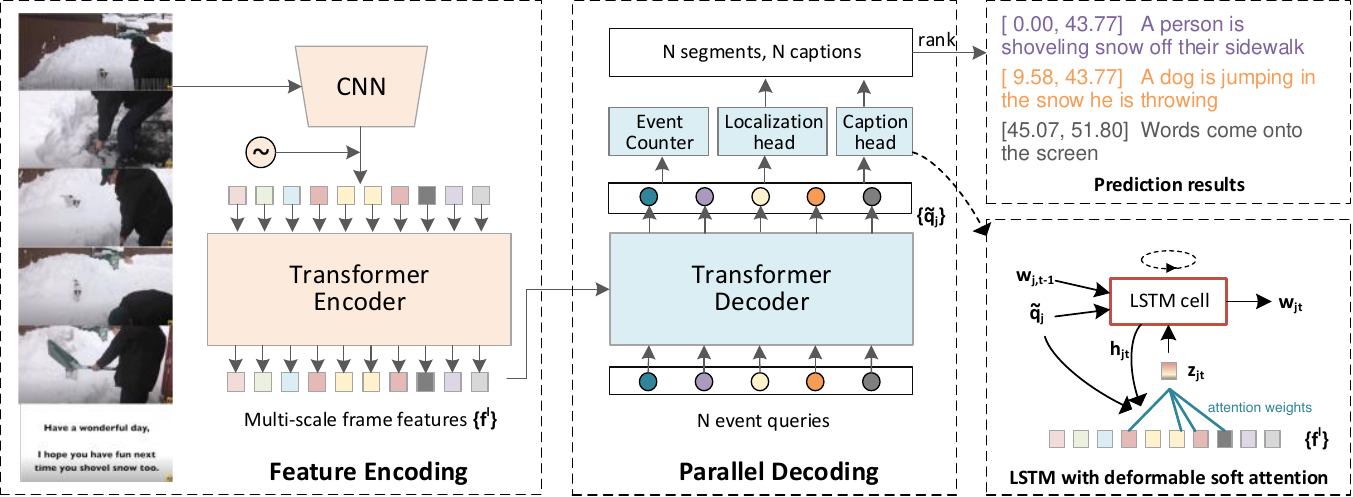}
    \caption{Overview of the proposed method. Firstly, we adopt a pre-trained video feature extractor and a transformer encoder to obtain a sequence of frame-level features. A transformer decoder and three prediction heads are then proposed to predict the locations, captions, and the number of events given learnable event queries. We provide two types of caption heads, which are based on vanilla LSTM and deformable soft-attention enhanced LSTM, respectively. In the testing stage, we select the top detected events by ranking the captioning score and localization score, with no need to remove redundancy by non-maximum suppression.}
    
    \label{fig:overview}
\end{figure*}

\vspace{0.5em}
\noindent{\textbf{Temporal event proposals.}}
Temporal event proposals~(TEP), also called temporal action proposals, aims to predict temporal segments containing event instances in untrimmed videos. Mainstream approaches can be divided into two categories: anchor-based and boundary-based. Anchor-based methods~\cite{heilbron2016fast, gao2017turn, shou2016temporal, escorcia2016daps} pre-define a vast number of anchors at different scales with regular intervals, followed by the evaluation network to score each anchor. However, the pre-defined scales and intervals can not cover all temporal patterns, especially in videos with variable temporal scales. The boundary-based methods~\cite{zhao2017temporal, lin2018bsn, lin2019bmn, lin2020fast} combine salient frames with high confidences to form proposals in a local-to-global manner. Both types of methods contain hand-crafted designs (\eg, NMS and rule-based label assignment), which require careful manual threshold selection and are not a strictly end-to-end method.

\vspace{0.5em}
\noindent{\textbf{Dense video captioning.}}
Dense video captioning is a multi-task problem that combines event localization and event captioning. Krishna et al.~\cite{krishna2017dense} propose the first dense video captioning model, containing a multi-scale proposal module for localization and an attention-based LSTM for context-aware caption generation. Some of the following work aims to enrich the event representations by context modeling~\cite{wang2018bidirectional, yang2018hierarchical}, event-level relationships~\cite{wang2020event}, or multi-modal feature fusion~\cite{Iashin2020MDVC, Iashin2020better}, enabling more accurate and informative captioning generation. 

One of the limitations of the above approaches is that the localization module can not benefit from the captioning module. Some researchers try to explore the interaction between two sub-tasks. Li et al.~\cite{li2018jointly} introduce a proxy task, \ie, predicting language rewards of generated sentences, as an additional optimization goal of the localization module. Zhou et al.~\cite{zhou2018end} propose a differential masking mechanism to link the gradient flow from captioning loss to proposals' boundaries, enabling a joint optimization of two tasks. We argue that neither a binary mask vector~\cite{zhou2018end} nor a scalar descriptiveness score~\cite{li2018jointly} carries enough informative gradients of linguistic cues to guide the internal feature representation in the proposal module during the back-propagation training. Instead, the proposed PDVC exploits the inter-task interactions by enforcing the two sub-tasks share the same intermediate features. Moreover, we adopt one-to-one matching between intermediate feature vectors and target event instances to obtain the discriminative features for captioning, significantly different from previous methods with a many-to-one anchor assignment strategy.

Another promising direction focuses on the coherence of generated captions. Early work~\cite{krishna2017dense, li2018jointly, zhou2018end, wang2018bidirectional} usually generates a large number of proposal-caption pairs (10 times more than the number of ground-truth events) for a high recall, where massive redundancy greatly reduces the readability and coherence of generated captions. SDVC~\cite{Mun2019stream} is the first to tackle this problem by introducing a ``localize-select-describe" pipeline. Given the output proposals produced by a TEP model, they develop an RNN-based event sequence generation network (ESGN) to select a small set of proposals, reducing the predicted proposal number from 100 to 2.85 on average. Though promising performance is achieved, SDVC is not an end-to-end model, making a multi-step training strategy necessary. The recurrent nature also restricts the application of ESGN to handle long videos with a large number of events. We parallelize localization, selection, and captioning tasks in a single end-to-end framework, largely simplifying the pipeline while enabling generating accurate and coherent captions.

\vspace{0.5em}
\noindent{\textbf{Transformer-based detector.}}
Transformer~\cite{vaswani2017attention} is an encoder-decoder architecture based on an attention mechanism for natural language processing. Benefit from the significant ability to capture long-range relationships,  Transformer has been successfully applied and shows promising performance in computer vision~\cite{dosovitskiy2020image, touvron2020training, girdhar2019video, yang2020learning, chen2020pre}. Detection Transformer (DETR)~\cite{carion2020end} is a newly emerging solution to object detection, which considers object detection as a set prediction task and does not rely on any hand-crafted components. Though it offers promising performance, DETR suffers from the high training time due to the slow convergence of global attention mechanism. Deformable Transformer~\cite{Zhu2020Deform} is proposed to speed up the network training and gain better performance by attending to sparse spatial locations of images and incorporating multi-scale feature representation. Inspired by the simple design and promising performance of DETR-style detectors in the image domain, we extend Deformable Transformer into a more challenging dense video captioning task in the video domain.

\section{Method}

To simplify the dense video captioning pipeline and explore the mutual benefits between localization task and captioning task, we directly detect a set of temporally-localized captions with an appropriate size $\{(t^s_j, t^e_j, S_j)\}_{j=1}^{N_{\rm set}}$, where $t^s_j$, $t^e_j$, $S_j$ represent the starting time, ending time, and the caption of an event, respectively. The set size $N_{\rm set}$ is also predicted by PDVC.

Specifically, a deformable transformer with an encoder-decoder structure is adopted to capture the inter-frame, inter-event, and event-frame interactions by attention mechanism and produce a set of event query features. Then, two parallel prediction heads predict the boundaries and captions for each event query simultaneously. An event counter predicts the event number $N_{\rm set}$ from a global view. The final results are obtained by select top $N_{\rm set}$ events with high confidence to ensure a complete and coherent story. Fig.~\ref{fig:overview} shows the overview of the proposed PDVC.

\subsection{Preliminary: Deformable Transformer}
Deformable Transformer~\cite{Zhu2020Deform} is an encoder-decoder architecture based on multi-scale deformable attention (MSDAtt). MSDAtt mitigates the slow convergency problem of the self-attention~\cite{vaswani2017attention} in Transformer when processing image feature maps, by attending to a sparse set of sampling points around reference points. Given multi-scale feature maps $\mathbf{X}=\{\mathbf{x}^l\}_{l=1}^{L}$ where $\mathbf{x}^l \in \mathbb{R}^{C\times H\times W}$, a query element $\mathbf{q}_j$ and a normalized reference point $\mathbf{p}_j\in[0, 1]^2$, MSDAtt outputs a context vector by the weighted sum of $K$$\times$$L$ sampling points across feature maps at $L$ scales:
\begin{equation}
  \begin{aligned}
    &{\text{MSDAtt}} (\mathbf{q}_j, \mathbf{p}_j, \mathbf{X}) = \sum_{l=1}^L \sum_{k=1}^K A_{jlk} \mathbf{W} \mathbf{x}^l_{\Tilde{\mathbf{p}}_{jlk}} \\
    &\Tilde{\mathbf{p}}_{jlk} = \phi_l(\mathbf{p}_j) + \Delta \mathbf{p}_{jkl},
  \end{aligned}
\label{eq2}
\end{equation}
where $\Tilde{\mathbf{p}}_{jkl}$ and $A_{jkl}$ are the position and the attention weight of $k$-th sampled key at $l$-th scale for $j$-th query element, respectively. $\mathbf{W}$ is the projection matrix for key elements. $\phi_l$ projects normalized reference points into the feature map at $l$-th level. $\Delta \mathbf{p}_{jkl}$ are sampling offsets w.r.t. $\phi_l(\mathbf{p}_j)$. Both $A_{jkl}$ and $\Delta \mathbf{p}_{jkl}$ are obtained by linear projection onto the query element. Note that the original MSDAtt applies multi-head attention mechanism, while here we illustrate the single-head version for better understanding. 

Deformable Transformer replaces the self-attention modules in the Transformer encoder and the cross-attention modules in the Transformer decoder with deformable attention modules, enabling a fast convergence rate and better representation ability in object detection.

\subsection{Feature Encoding}
To capture rich spatio-temporal features in a video, we first adopt a pre-trained action recognition network (\eg, C3D~\cite{tran2015learning}, TSN~\cite{wang2018temporal}) to extract the frame-level features. We re-scale the feature map's temporal dimension to a fixed number $T$ by interpolation to facilitate batch processing. Then, to better utilize multi-scale features for predicting multi-scale events,  we add $L$ temporal convolutional layers (stride=2, kernel size=3) to get feature sequences across multiple resolutions, from $T$ to $T/2^{L}$. The multi-scale frame features with their positional embedding~\cite{vaswani2017attention} are fed into the deformable transformer encoder, extracting the frame-frame relationship across multiple scales. The output frame features are denoted as $\{\mathbf{f}^l\}_{l=1}^L$.

\subsection{Parallel Decoding}

The decoding network contains a deformable transformer decoder and three parallel heads, a captioning head for caption generation, a localization head to predict events' boundaries with confidence scores, and an event counter to predict a suitable event number. The decoder aims to directly query the event-level features from frame features conditioned on $N$ learnable embedding (termed event queries) $\{\mathbf{q}_j\}_{j=1}^N$, and their corresponding scalar referent points ${p}_j$. Note that ${p}_j$ is predicted by a linear projection with a sigmoid activation over ${\mathbf{q}}_j$. The event queries and reference points serve as the initial guess of the events' features and locations (center points), and they will be refined iteratively at each decoding layer, as in~\cite{Zhu2020Deform}. The output query features and the reference point are denoted as ${\tilde{{\mathbf{q}}}_j}, {\tilde{{\mathbf{p}}}_j}$.

\vspace{0.5em}
\noindent{\textbf{Localization head.}} Localization head performs box prediction and binary classification for each event query. Box prediction aims to predict the 2D relative offsets (center and length) of the ground-truth segment w.r.t. the reference point. Binary classification aims to generate the foreground confidence of each event query. Both box prediction and binary classification are implemented by multi-layer perceptron. After that, we obtain a set of tuples $\{t^{\rm s}_j, t^{\rm e}_j, c^{\rm loc}_j\}_{j=1}^{N}$ to represent the detected events, where $c^{\rm loc}_j$ is the localization confidence of the event query ${\tilde{{\mathbf{q}}}_j}$.

\vspace{0.5em}
\noindent{\textbf{Captioning head.}}
We provide two captioning heads, a lightweight one and a standard one. The lightweight head simply feeds ${\tilde{{\mathbf{q}}}_j}$ into a vanilla LSTM at each timestamp. The word $w_{jt}$ is predicted by an FC layer followed by a softmax activation over the hidden state $\mathbf{h}_{jt}$ of LSTM.

However, the lightweight captioning head only receives an event-level representation ${\tilde{{\mathbf{q}}}_j}$, lacking the interactions between linguistic cues and frame features. Soft attention~(SA)~\cite{yao2015describing, wang2018bidirectional, Mun2019stream} is a widely-used module in video captioning, which can dynamically determine the importance of each frame when generating a word. Traditional two-stage methods~\cite{wang2018bidirectional, Mun2019stream} aligns the event segments and their captions by restricting the attention area being within the event boundaries, but our captioning head can not access events' boundaries, increasing the optimization difficulty to learn relationships between the linguistic word and frames. To alleviate this problem, we propose the deformable soft attention (DSA) to enforce the soft attention weights focus on a small area around the reference points.  Specifically, when generating the $t$-th word $w_t$, we first generate $K$ sampling points from each $\mathbf{f}^l$ conditioned on both language query $\mathbf{h}_{jt}$ and event query $\mathbf{\tilde{q}}_j$, following Eqn.~\ref{eq2}, where $\mathbf{h}_{jt}$ denotes the hidden state in LSTM. Then we consider $K$$\times$$L$ sampling points as the key/value and $[\mathbf{h}_{jt}, \mathbf{\tilde{q}}_j]$ as the query in soft attention. Since the sampling points are distributed around the reference point ${\tilde{{\mathbf{p}}}_j}$, the output features $\mathbf{z}_{jt}$ of DSA are restricted to attend on a relatively small region. The LSTM takes as input the concatenation of the context features $\mathbf{z}_{jt}$, event query features $\mathbf{\tilde{q}}_j$ and previous words $w_{j,t-1}$.
The probability for next word $w_{jt}$ is obtained by an FC layer with softmax activation over $\mathbf{h}_{jt}$. With the evolving of LSTM, we obtain a sentence $S_j=\{w_{j1}, ..., w_{jM_j}\}$, where $M_j$ is the sentence length.

\vspace{0.5em}
\noindent{\textbf{Event counter.}}
Considering that an appropriate event number is an essential indicator for dense captioning quality: too many events cause replicated captions and bad readability; too few detected events mean information missing and incomplete story. The event counter aims to detect the event number of the video. It contains a max-pooling layer and an FC layer with softmax activation, which first compress the most salient information of event queries $\{\mathbf{\tilde{q}}_j\}$ to a global feature vector, and then predict a fix-size vector $\mathbf{r}_{\rm len}$, where each value refers to the possibility of a specific number. During the inference stage, predicted event number is obtained by $N_{\rm set} = {\rm argmax}(\mathbf{r}_{\rm len})$. The final outputs are obtained by selecting the top $N_{\rm set}$ events with accurate boundaries and good captions from $N$ event queries. The confidence of each event query is calculated by:
\begin{equation}
c_{j} = c^{\rm loc}_j + \frac{\mu}{{M_j}^{\gamma}} \sum_{t=1}^{M_j} \log(c^{\rm cap}_{jt}),   \label{rank}
\end{equation}
where $c^{\rm cap}_{jt}$ is the probability of the generated word. We observe that the average word confidence is not a convincing measurement of sentence-level confidence since the captioning head tends to produce overestimated confidence for short sentences. Thus, we add a modulation factor $\gamma$ to rectify the influence of caption length. $\mu$ is the balance factor.

\vspace{0.5em}
\noindent{\textbf{Set prediction loss.}}
During training, PDVC produces a set of $N$ events with their locations and captions. To match predicted events with ground truths in a global scheme, we use the Hungarian algorithm following~\cite{carion2020end} to find the best bipartite matching results. The matching cost is defined as $C = \alpha_{\rm giou} L_{\rm giou} + \alpha_{\rm cls} L_{\rm cls}$, 
where $L_{\rm giou}$ represents the generalized IOU~\cite{rezatofighi2019generalized} between predicted temporal segments and ground-truth segments, $L_{\rm cls}$ represents the focal loss~\cite{lin2017focal} between the predicted classification score and the ground-truth label. 
The matched pairs are selected to calculate the set prediction loss, which is the weighted sum of gIOU loss, classification loss, countering loss, and caption loss:
\begin{equation}
    L = \beta_{\rm giou} L_{\rm giou} + \beta_{\rm cls} L_{\rm cls} + \beta_{\rm ec} L_{\rm ec} + \beta_{\rm cap} L_{\rm cap},
\end{equation}
where $L_{\rm cap}$ measures the cross-entropy between the predicted word probability and the ground truth normalized by the caption length, $L_{\rm ec}$ is also the cross-entropy loss between the predicted count distribution and the ground truth.  

Note that we follow~\cite{carion2020end, Zhu2020Deform} to add prediction heads to each layer of the transformer decoder. The final loss is the summation of the set prediction losses of all decoder layers.

\vspace{0.5em}
\noindent{\textbf{PDVC for paragraph captioning.}}
Paragraph captioning~\cite{park2019adver,lei2020mart,yu2016video} is a simplified version of dense video captioning, which focuses on generating a coherent paragraph and does not need to predict the temporal location of each sentence. PDVC can easily extend to paragraph captioning by removing the localization function and taking the pre-extracted proposals as input event queries. Specifically, we consider the linear embeddings of proposals' position as the event queries and use the proposals' center as reference points. Then PDVC is trained with caption loss only. 

\begin{table*}[h]
\small
\begin{minipage}{\textwidth}
 \begin{minipage}[t]{0.6\textwidth}
  \centering
     \makeatletter\def\@captype{table}\makeatother
       \setlength{\tabcolsep}{0.75 mm}{
        \begin{tabular}{c |ccccc| ccccc|c}
            \toprule
            \multirow{2}{*}{Method}&
            \multicolumn{5}{c|}{Recall}&\multicolumn{5}{c|}{Precision} & \multirow{2}{*}{F1} \\
             &0.3&0.5&0.7&0.9 &avg &0.3&0.5&0.7&0.9&avg&  \\
            \midrule
            MFT~\cite{Xiong2018ECCV}  & 46.18 & 29.76 & 15.54 & 5.77 & 24.31 & 86.34 & 68.79 & 38.30 & 12.19 & 51.41 & 33.01 \\
            SDVC~\cite{Mun2019stream}  & \textbf{93.41} & \textbf{76.40} & 42.40 & 10.10 & {55.58} & 96.71 & 77.73 & \textbf{44.84} & 10.99 & 57.57 & 56.56 \\
            \textbf{PDVC\_light} & 88.78 & 71.74 & \textbf{45.70} & \textbf{17.45} & \textbf{55.92}& 96.83 & 78.01 & 41.05 & \textbf{14.69} & 57.65 & \textbf{56.77}\\
            \textbf{PDVC} &  89.47 & 71.91 & {44.63} & {15.67} & 55.42 & \textbf{97.16} & \textbf{78.09} & 42.68 & 14.40 & \textbf{58.07} & {56.71} \\
            \bottomrule
        \end{tabular}}
        \caption{Event localization on the ActivityNet Captions validation set.}
        \label{table:EventLoc}
\end{minipage}
\begin{minipage}[t]{0.08\textwidth}
~
\end{minipage}
\begin{minipage}[t]{0.30\textwidth}
   \centering
        \makeatletter\def\@captype{table}\makeatother
        \setlength{\tabcolsep}{0.8 mm}{
        \begin{tabular}{lcccc}
            \toprule
            \multirow{2}{*}{Method}  & \multicolumn{4}{c}{{Predicted proposals}} \\
            & \multirow{1}{*}{B4} & \multirow{1}{*}{M}  & \multirow{1}{*}{C} & \multirow{1}{*}{SODA\_c} \\
            \midrule
            MT~\cite{zhou2018end} &0.30 & 3.18  & 6.10 & - \\
            ECHR~\cite{wang2020event} & - & 3.82 & - & - \\
            \textbf{PDVC\_light} & \textbf{{0.89}}  & {4.56}  & \textbf{{23.07}} & 4.34 \\
            \textbf{PDVC} & 0.80 & \textbf{4.74} & 22.71  & \textbf{4.42} \\
            \bottomrule
        \end{tabular}}
        \caption{Dense captioning on YouCook2.}
        \label{table:SotaYC}
   \end{minipage}
\end{minipage}
\end{table*}

\begin{table*}[h]
\small
\begin{minipage}{\textwidth}
 \begin{minipage}[t]{0.61\textwidth}
  \centering
     \makeatletter\def\@captype{table}\makeatother
     \setlength{\tabcolsep}{0.9 mm}{
\begin{tabular}{l|c|ccc|cccc}
    \toprule
    \multirow{2}{*}{Method} & \multirow{2}{*}{Features} &\multicolumn{3}{c|}{{Ground-truth proposals}} & \multicolumn{4}{c}{{Predicted proposals}} \\
    &  & {~~~B4~~~} & {~~~M~~~} & {~~~C~~~} & {B4} & {~M~}  & {~C~} & SODA\_c \\
    \midrule
    DCE~\cite{krishna2017dense} & C3D & 1.60 & 8.88 & 25.12 & 0.17 & 5.69 & 12.43& - \\
    TDA-CG~\cite{wang2018bidirectional}{$^*$} & C3D & - & {9.69} & - & 1.31 & 5.86 & 7.99 & - \\
    DVC~\cite{li2018jointly} & C3D& 1.62 & 10.33 & 25.24 & 0.73 & 6.93 & 12.61 & - \\
    SDVC~\cite{Mun2019stream} &C3D& - & - & - & - & 6.92 & - & -\\
    Efficient~\cite{Suin2020efficient} & C3D & - & - & - & 1.35 & 6.21 & 13.82 & -  \\
    ECHR~\cite{wang2020event} & C3D& 1.96 & \textbf{10.58} & 39.73 & 1.29 & 7.19 & 14.71 & 3.22 \\
    \textbf{PDVC\_light} & C3D& 2.61 & 10.48 & \textbf{47.83} & 1.51 & 7.11 & \textbf{26.21} & 5.17 \\ 
    \textbf{PDVC} & C3D& \textbf{2.64} & 10.54 & 47.26 & \textbf{1.65} & \textbf{7.50} & {25.87} & \textbf{5.26} \\
    \midrule
    MT~\cite{zhou2018end}{$^*$} & TSN & 2.71 & 11.16 & 47.71 & 1.15 & 4.98 & 9.25 & - \\
    \textbf{PDVC} & TSN & \textbf{3.07} & \textbf{11.27} & \textbf{52.53} & \textbf{1.78} & \textbf{7.96} & \textbf{28.96} & \textbf{5.44} \\
    \midrule
    MDVC~\cite{Iashin2020MDVC}{$^{*\dagger}$} & I3D+VGGish & 1.98 & 11.07 & 42.67 & 1.01 & 6.86 & 7.77 & - \\
    BMT~\cite{Iashin2020better}{$^{*\dagger}$} & I3D+VGGish & 1.99 & 10.90 & 41.85 & 1.88 & 7.43 & 11.94& - \\
    \textbf{PDVC}{$^{\dagger}$} & I3D+VGGish & \textbf{3.12} & \textbf{11.26} & \textbf{53.65} & \textbf{1.96} & \textbf{8.08} & \textbf{28.59} & \textbf{5.42} \\
    \bottomrule
\end{tabular}}
\caption{Dense captioning on the ActivityNet Captions validation set. B4/M/C is short for BLEU4/METEOR/CIDEr. $^*$ indicates results re-evaluated by the same evaluation toolkit. $^{\dagger}$ means results with part of the dataset (9\% videos missing). }
\label{table:SotaANET}
\end{minipage}
\begin{minipage}[t]{0.04\textwidth}
~
\end{minipage}
\begin{minipage}[t]{0.34\textwidth}
  \centering
        \makeatletter\def\@captype{table}\makeatother
        \setlength{\tabcolsep}{0.7 mm}{
\begin{tabular}{lcccc}
    \toprule
    \multirow{2}{*}{Method}  & \multirow{2}{*}{Features} & \multirow{2}{*}{B4} & \multirow{2}{*}{M}  & \multirow{2}{*}{C} \\
    \\
    \midrule
    \multicolumn{5}{l}{\textbf{Ground-truth proposals}} \\
    HSE~\cite{Zhang2018ECCV} & V &9.84 & 13.78 & 18.78  \\
    MART~\cite{lei2020mart} & V+F & 10.33 & 15.68 & 23.42 \\
    VTrans~\cite{zhou2018end} & V+F & 9.75 & 15.64 & 22.16 \\
    Trans-XL~\cite{Dai2019trans} & V+F & 10.39 & 15.09 & 21.67 \\
    GVD~\cite{Zhou2019ground} & V+F+O & 11.04 & 15.71 & 21.95 \\
    GVDsup~\cite{Zhou2019ground}& V+F+O & 11.30 & 16.41 & 22.94 \\ 
    AdvInf~\cite{park2019adver} & V+F+O & 10.04 & \textbf{16.60} & 20.97 \\
    \textbf{PDVC} & V+F & \textbf{11.80} & 15.93  & \textbf{27.27} \\
    \midrule
    \multicolumn{5}{l}{\textbf{Predicted proposals}} \\
    MFT~\cite{Xiong2018ECCV} & V+F &\textbf{10.29} & 14.73  & 19.12 \\
    \textbf{PDVC} & V+F & {10.24} & \textbf{15.80}  & \textbf{20.45} \\
    \bottomrule
\end{tabular}}
\caption{Paragraph captioning on the ActivityNet Captions \textit{ae-val} set~\cite{Zhou2019ground}. V/F/O refers to visual/flow/object features.}
\label{table:SotaParaCap}
\end{minipage}
\end{minipage}
\vspace{-1em}
\end{table*}

\section{Experiments}

\subsection{Experimental Settings}

\paragraph{Datasets.}
We use two large-scale benchmark datasets, ActivityNet Captions~\cite{krishna2017dense}, and YouCook2~\cite{zhou2018towards} to evaluate the effectiveness of the proposed PDVC. ActivityNet Captions contains 20k long untrimmed videos of various human activities. On average, each video lasts 120s and is annotated with 3.65 temporally-localized sentences. We follow the standard split with 10009/4925/5044 videos for training, validation, and test. YouCook2 has 2000 untrimmed videos of cooking procedures with an average duration of 320s. Each video has 7.7 annotated segments with associated sentences. We use the official split with 1333/457/210 videos for training, validation, and test. 

\vspace{0.5em}
\noindent{\textbf{Evaluation metrics.}}
We evaluate our method in three aspects:  1) For localization performance, we use the average precision, average recall across IOU at \{0.3, 0.5, 0.7, 0.9\}  and their harmonic mean, F1 score. 2)  For dense captioning performance, we follow the \href{https://github.com/ranjaykrishna/densevid_eval/tree/deba7d7e83012b218a4df888f6c971e21cfeea33}{official evaluation tool} provided by ActivityNet Challenge 2018, which calculates the average precision (measured by BLEU4~\cite{papineni2002bleu}, METEOR~\cite{lavie2005meteor}, and CIDEr~\cite{vedantam2015cider}) of the matched pairs between generated captions and the ground truth across IOU thresholds of \{0.3, 0.5, 0.7, 0.9\}. However, the official scorer does not consider the storytelling quality, \ie, how well the generated captions can cover the video's whole story. We further adopt \href{https://github.com/fujiso/SODA/tree/22671b3570e088217139bcb1e4de7a3499c30294}{SODA\_c}~\cite{fujitasoda} for an overall evaluation. 3) For paragraph captioning performance, we form a paragraph by sorting generated captions according to their starting time and report the \href{https://github.com/jayleicn/recurrent-transformer/tree/3c31d2444c178ccbc78998d2ae1d4910b02b95ae/densevid_eval}{paragraph-level captioning performance}.
Note that ActivityNet Captions has two sets of annotations for the validation set. For SODA\_c, we evaluate it by two sets independently and report their average score.

\vspace{0.5em}
\noindent{\textbf{Implementation details.}}
For ActivityNet Captions, we use a C3D~\cite{tran2015learning} pre-trained on Sports1M~\cite{karpathy2014large} to extract frame-level features. To fairly compare with state-of-the-art methods, we also test our model based on TSN~\cite{wang2018temporal} features provided by~\cite{zhou2018end}, and I3D+VGGish features provided by~\cite{Iashin2020better}. For YouCook2, we use the same TSN features as in~\cite{zhou2018end}. 

We use a two-layer deformable transformer with multi-scale (4 levels) deformable attention. The deformable transformer uses a hidden size of 512 in MSDAtt layers and 2048 in feed-forward layers. The number of event queries is 10/100 for ActivityNet Captions/YouCook2. We implement a lightweight PDVC (termed PDVC\_light) with the vanilla LSTM captioner and the standard PDVC with the LSTM-DSA captioner. The LSTM hidden dimension in captioning heads is 512. For the event counter, we choose the maximum count as 10/20 for ActivityNet Captions/YouCook2. In Eqn.~\ref{rank}, the length modulation factor $\gamma$ is set to 2, and the trade-off ratio $\mu$ is set to 0.3/1.0 for PDVC\_light/PDVC. The cost ratios in the bipartite matching are $\alpha_{\rm giou}$:$\alpha_{\rm cls}$=2:1 and the loss ratios are $\beta_{\rm giou}$:$\beta_{\rm cls}$:$\beta_{\rm ec}$:$\beta_{\rm cap}$=2:1:1:1. We use Adam~\cite{kingma2015adam} optimizer with an initial learning rate of 5e-5 and the mini-batch size of 1 video.

\subsection{Comparison with State-of-the-art Methods}

\vspace{0.5em}
\noindent{\textbf{Localization performance.}} The comparison of event localization quality is shown in Table~\ref{table:EventLoc}. SDVC and MFT generate event proposals by a sophisticated ``localize-select-describe" workflow. 
In contrast, PDVC removes the hand-crafted designs in the traditional proposal modules and directly outputs the proposals in a parallel manner, which is more efficient to deal with long sequences than recurrent counterparts. We surpass MFT by a large margin and achieve similar (slightly better) performance to SDVC, which shows the effectiveness of parallel set prediction in our method. Besides, the choice of the captioning head can slightly influence the balance of precision and recall. 

\vspace{0.5em}
\noindent{\textbf{Dense captioning performance.}}
In Table~\ref{table:SotaANET}, we list the performance of state-of-the-art models with cross-entropy training\footnote{{A few methods~\cite{Mun2019stream, wang2020event} incorporates Reinforcement Learning (RL)~\cite{Rennie2017self} after the cross-entropy training to further boost the performance. Note that we do not compare with these methods since RL training requires a more complex captioning network (\eg, Hierarchical RNN~\cite{yu2016video}) and extra-long training time, which is opposite to the design philosophy of PDVC. Moreover, RL training tends to produce longer sentences with repeated phrase~\cite{wang2019describe}, lowering the coherence and readability of generated captions.}}
on ActivityNet Captions. With ground-truth proposals, PDVC achieves considerable improvement over the state-of-the-art on BLEU4 and CIDEr, which shows a deformable transformer plus an LSTM captioner can give good caption quality. With predicted proposals, PDVC with C3D features achieves the best performance on BLEU4/METEOR/CIDEr/SODA\_c, giving a 22.22\%/4.31\%/75.87\%/63.35\% relative improvement over state-of-the-art scores. We find that PDVC with ground-truth proposals does not show much superiority over ECHR on METEOR but surpasses ECHR with predicted proposals, indicating the generated proposals of PDVC are much better.  Even with a lightweight LSTM as a captioner, PDVC\_light can surpass most two-stage approaches on BLEU4/CIDEr/SODA\_c. The reason mainly comes from the parallel decoding of the captioning head and localization head, which helps to generate proposals with high descriptiveness and discriminative internal representation. 

Table~\ref{table:SotaYC} shows the dense captioning performance on the YouCook2 validation set. Our method achieve state-of-the-art performance with a considerable performance gain over other methods on all metrics.

\vspace{0.5em}
\noindent{\textbf{Paragraph captioning performance.}}
Table~\ref{table:SotaParaCap} shows the comparison between PDVC and state-of-the-art paragraph captioning methods. With ground-truth proposals, 
PDVC with a deformable transformer plus an attention-based LSTM can surpass several transformer-based captioning models, like MART, VTrans, and Trans-XL, indicating the strong representation ability of deformable attention in the encoder-decoder and the LSTM-DSA. It is promising for PDVC to get a further performance boost by incorporating a transformer captioner.  We leave this for future work.

Even with predicted proposals, we observe PDVC has a comparable performance with previous methods with ground-truth proposals, indicating that query features contain rich information covering main segments in videos. While most previous paragraph captioning methods require ground-truth annotation at testing, our model reduces the captioning module's dependence on accurate proposals by parallel decoding, raising the possibility of generating good paragraphs without human annotations of the test video.

\vspace{0.5em}
\noindent{\textbf{Efficiency.}} We compare the inference time of PDVC against two-stage methods~TDA-CG~\cite{wang2018bidirectional}, MT~\cite{zhou2018end} under the same hardware environment in Table~\ref{table:eff}. Our methods are more efficient since: 1) Only a few event proposals with their captions are predicted in parallel; 2) We do not require a dense-to-sparse proposal selection like NMS; 3) MSDAtt is an efficient operator due to the sparse sampling. 
%

\subsection{Interaction between Localization \& Captioning}
\label{sec: interaction}

In this part, we go deeper into the mutual effect between two subtasks. It is straightforward that localization can aid captioning since the localization supervision guides the query features to specific ground-truth regions, which contains rich semantics matching the target captions. Therefore, we focus on how captioning task affects proposals' quality, which is less explored in the previous literature.

\vspace{0.5em}
\noindent{\textbf{Captioning supervision helps generate proposals with descriptiveness.}} To better study the quality of proposals generated by PDVC,  
we use the same pre-trained event captioning model~\cite{yao2015describing} to evaluate the descriptiveness of generated proposals of different models. 
We also reimplemented two mainstream proposal generation modules SST and SST+ESGN for comparison. 
Both SST and SST+ESGN are trained with localization loss only, while PDVC is trained with both localization and captioning loss. As shown in Table~\ref{table:LossType}, PDVC achieves a slightly lower F1 score but the best descriptiveness score among the four models.

\begin{figure}[t]
    \centering
    \includegraphics[width= 0.99 \columnwidth]{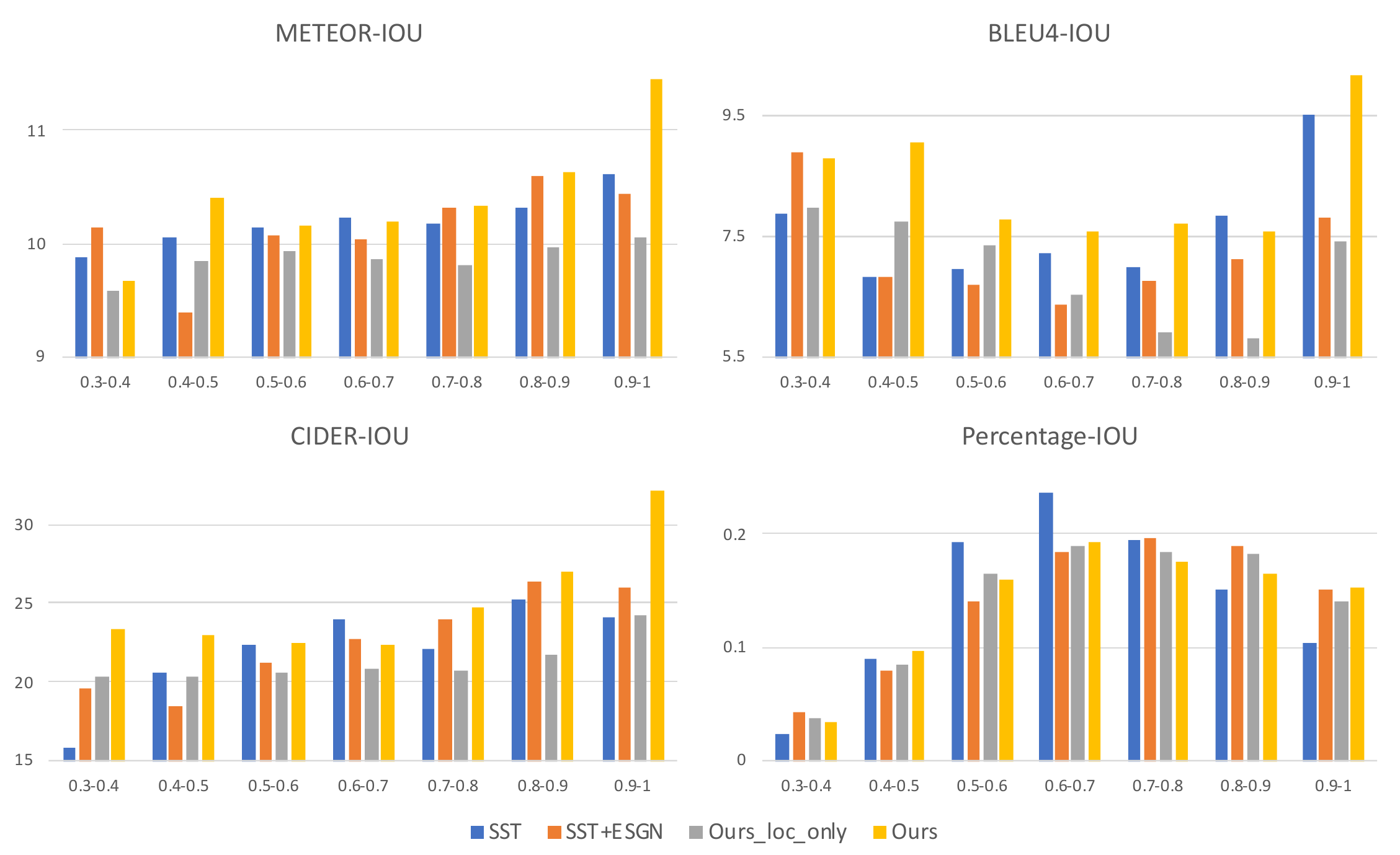}
\caption{Caption quality vs. IOU.  We omit the pairs with IOU$<0.3$ (less than 2\% of all pairs).}
\label{fig:Interact} 
\end{figure}

We match each generated proposal with one ground-truth segment with the highest overlap. Fig.~\ref{fig:Interact} demonstrates the statistics of matching results. Surprisingly, incorporating caption supervision yields a considerable boost in caption quality of high-precision proposals (\ie, IOU$>$0.9). The reason may be that the captioning head is trained based on event query features corresponding to accurate proposals, so PDVC learns to enhance the descriptiveness of the high-precision proposals. The last subfigure shows the IOU distribution of matched pairs. Most proposals produced by SST are not very accurate (mainly with 0.5$<${\rm IOU}$<$0.8). When further incorporating ESGN for adaptively proposal selection, the majority of proposals are with 0.6$<${\rm IOU}$<$0.9. Ours and Ours\_loc\_only achieve a similar IOU distribution to SST+ESGN, but do not introduce any hand-crafted components like anchor generation and NMS.

\begin{table}[]
\setlength{\tabcolsep}{1.3mm}{
\small
\begin{tabular*}{8.3cm}{l c c c c c}
    \toprule
    Method &  BAF-CG~[24] & MT~[31]  & \textbf{PDVC\_light} & \textbf{PDVC}\\
    \midrule
    Time(secs) & 2.39 & 2.05 & \textbf{0.09} & \textbf{0.16}  \\
    \bottomrule
\end{tabular*}}
\caption{Inference speed. We report average inference time (secs/video) of 100 sampled videos with a single Tesla V100 GPU.}
\vspace{-0.5 em}
\label{table:eff}
\end{table}

\begin{table}[]
\fontsize{8.5pt}{1em} \selectfont
    \centering
        \setlength{\tabcolsep}{0.6 mm}{
        \begin{tabular}{l c ccc cccc}
            \toprule
            {Method} &{Loss} &{\#p} & {Rec.} &{Pre.} & {F1} & {B4} & {M}  &{C} \\
            \midrule
            {SST~\cite{buch2017sst}} & loc. & 3.00 & 42.00 & 60.99 &49.74 & 0.98 & 6.70 & 17.34  \\
            {SST+ESGN~\cite{Mun2019stream}} &loc.& 2.79 & 53.80 & 61.37 & 57.33 & 1.09 & 6.80 & 19.67 \\
            {Ours\_loc\_only} &loc. & 3.26 & {56.35} & 58.69 & {57.49} & 0.98 & 6.71 & 19.36 \\
            \textbf{Ours (PDVC)} & loc.+cap. & 3.03 & 55.42 & 58.07 & 56.71 & \textbf{1.24} & \textbf{7.03} & \textbf{21.91}\\
            \bottomrule
        \end{tabular}}
        \caption{Proposal quality with different loss types. Rec./Pre./F1 measures the localization performance, while B4/M/C measures dense captioning performance. \#p is the number of proposals.}
        \label{table:LossType}
        \vspace{-0.5em}
\end{table}

\begin{figure*}
    \centering
    \includegraphics[width=1. \textwidth]{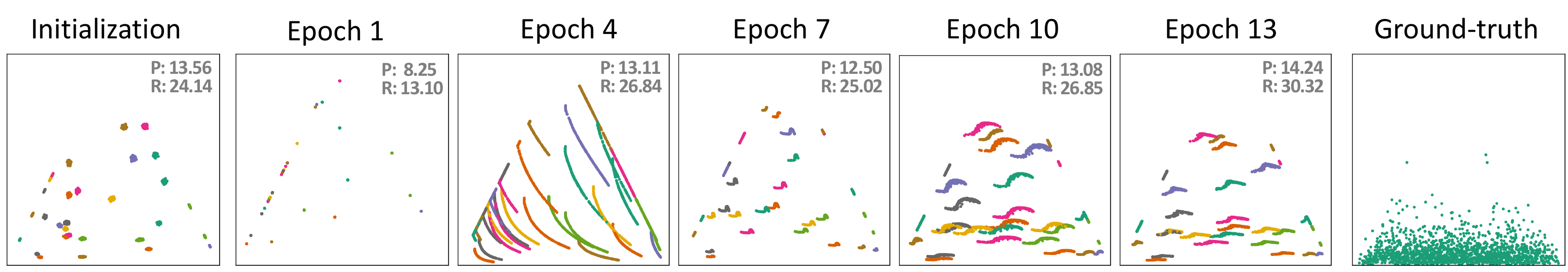}
    \caption{The distribution of predicted proposals without localization supervision. We plot the predicted proposals of 200 randomly sampled videos in the YouCook2 validation set. Horizontal and vertical axes represent the re-scaled center position and re-scaled length of proposals, respectively. Each sub-figure contains 30 clusters with different colors corresponds to 30 input event queries. R and P refer to recall and precision of the 30 generated proposals, respectively.}
    \label{fig:weak}
\end{figure*}

\begin{table*}[h]
\small
    \begin{subtable}[h]{0.38\textwidth}
        \centering
        \setlength{\tabcolsep}{0.65 mm}{
        \begin{tabular}{c c | c c c | c c}
                \toprule
                \multicolumn{2}{c|}{Transformer} & \multicolumn{3}{c|}{Captioning head} & \multirow{2}{*}{M} & \multirow{2}{*}{SODA\_c}  \\
                Vanilla & Deformable & LSTM & SA & DSA &  &  \\
                \midrule
                $\surd$ & &$\surd$ & & & 6.10 & 3.06 \\
                & $\surd$& $\surd$ && &  7.11  & 5.17 \\
                & $\surd$ &$\surd$ & $\surd$ & & 6.15 & 3.40 \\
                & $\surd$ &$\surd$ & & $\surd$ & \textbf{7.50}  & \textbf{5.26}\\
                \bottomrule
            \end{tabular}}
            \caption{Ablating the deformable operations}
            \label{table:abl}
    \end{subtable}
    {~}
    \begin{subtable}[h]{0.35\textwidth}
        \centering
        \setlength{\tabcolsep}{0.8 mm}{
        \begin{tabular}{cccccc}
            \toprule
            \#q & counter & Rec. & Pre. & {M}  & SODA\_c\\
            \midrule
            5 & $\surd$ & 57.46 & 57.10 & 6.96 & 5.02 \\
            10 & $\surd$ &55.92 & 57.65 & 7.11 & 5.17 \\
            30 &$\surd$ & 53.35 & 59.08 & 7.18 & 4.90 \\
            100 &$\surd$ &51.88 &59.27 & 7.33 & 4.59 \\
            10 & $\times$ & 77.67 & 44.88 & 6.62 & 4.30 \\
            \bottomrule
        \end{tabular}}
        \caption{Varying query number \& event counter}
        \label{table:query}
    \end{subtable}
    {~}
    \begin{subtable}[h]{0.2\textwidth}
        \centering
        \setlength{\tabcolsep}{0.8 mm}{
        \begin{tabular}{c c c c c}
                \toprule
                \multirow{1}{*}{$\gamma$} & \multirow{1}{*}{Rec.} & \multirow{1}{*}{Pre.} & \multirow{1}{*}{M} & \multirow{1}{*}{SODA\_c}\\
                \midrule
                0.0 & 48.67 & 47.35 & 5.78 & 5.23  \\
                0.5 & 50.63 & 50.08 & 6.59 & 5.25 \\
                1.0 & 51.52 & 53.31 & 7.35 & 5.02 \\ 
                {2.0} & 55.92 & 57.65 & 7.11 & {5.17} \\
                3.0  & 55.72 & 57.87 & 6.90 & 5.19 \\
                \bottomrule
            \end{tabular}}
            \caption{Varying $\gamma$}
            \label{table:gamma}
    \end{subtable}
     \caption{Ablation studies on the ActivityNet Captions validation set. Subfigure (b) and (c) are based on PDVC\_light.}
     \vspace{-1em}
     \label{tab:temps}
\end{table*}

Generally speaking, descriptiveness is positively correlated to the precision of proposals with an ideal captioner. However, the performance of existing captioners is still far from satisfactory, which means they generate wrong or boring captions for some proposals. To reduce improper captions of the final results, it is essential to generate not only location-accurate but caption-aware proposals. Our model provides an effective solution to explore the mutual benefits between localization and captioning by parallel decoding.

\vspace{0.5em}
\noindent{\textbf{Captioning supervision helps learn location-aware features.}} Another advantage of parallel decoding is that we can directly remove the localization head to study the behavior of captioning head. We train an event proposal generation module based on merely captioning supervision, by making some modifications to the original PDVC to stabilize training, such as fixing the sampling offsets in the decoder and using the captioning cost in bipartite matching. More details can be found in the supplementary material. After iterative refinement in the decoder, we directly regard the reference points corresponding to event queries in the last decoder layer as event proposals. Fig.~\ref{fig:weak} shows the position distribution of predicted proposals on YouCook2. We also report quantitative results such as recall and precision.

As the training epoch increases, proposals' center tends to spread uniformly, and the proposals' length tends to focus on a relatively small value. Though a noticeable gap exists between the distributions of predicted proposals and ground-truth proposals, we see that the predicted proposals are gradually approaching the ground truth during training. The recall/precision at epoch 13 is 30.32/14.24, which is better than that at initialization (24.14/13.56). Based on the above findings, we argue that our method can implicitly capture the location-aware features from caption supervision, helping the optimization of the event localization.

\subsection{Ablation Studies}

\noindent{\textbf{Deformable components.}} As shown in Table~\ref{table:abl}, when removing deformable operations from deformable transformer or LSTM-DSA, the performance degrades considerably. We conclude that: 1) Adding locality into transformer helps to extract temporally-sensitive features for localization-aware tasks; 2) Focusing on a small segment around the proposals rather than the whole video helps the optimization of the event captioning.

\vspace{0.5em}
\noindent{\textbf{Query number \& event counter.}} As shown in Table~\ref{table:query}, only a few queries are sufficient for good performance. Too many queries lead to high precision and  METEOR, but low Recall and SODA\_c. We choose an appropriate query number for striking a balance of recall and precision. The final event number also controls the balance of precision and recall. The event counter can predict a reasonable number of event instances, making the generated captions reveal a whole story in the video.

\vspace{0.5em}
\noindent{\textbf{Length modulation.}} Table~\ref{table:gamma} shows that modulating the caption length~($\gamma$$>$$1$) obtains a better trade-off between METEOR\ \& SODA\_c and Precision \& Recall than averaging ($\gamma$$=$$1$) or summing ($\gamma$$=$$0$) the word scores.

\vspace{-0.2em}
\section{Conclusion}

\vspace{-0.1em}
This paper presents PDVC, an end-to-end dense video captioning framework with parallel decoding, which formulates dense video captioning as a set prediction task. PDVC directly produces a set of temporally-localized sentences without a dense-to-sparse proposal generation and selection process, significantly simplifying the traditional ``localize-then-describe" pipeline. Prediction heads for event localization and event captioning run in parallel to exploit the inter-task mutual benefits. Experiments on two benchmark datasets show that PDVC can generate high-quality captions and surpass state-of-the-art methods.

\noindent{\textbf{Acknowledgements.}}
This work was supported by the National Natural Science Foundation of China No.~61972188, 61903178, 61906081, and U20A20306, the General Research Fund of Hong Kong No.~27208720, and the Program for Guangdong Introducing Innovative and Enterpreneurial Teams No. 2017ZT07X386.

\clearpage
{\small
	\bibliographystyle{ieee}
	
}
\clearpage
\section{Supplementary Materials}
\subsection{More Implementation Details}

\vspace{0.5em}
\noindent{\textbf{Event proposal generation module based on merely captioning supervision.}}
In Sec.~\ref{sec: interaction}, we make the following modifications to train an event proposal generation module without localization supervision: 
1) We extend the 1D reference point to the 2D reference point $p_j = (p^{\rm c}_j, p^{\rm l}_j)$, where $p^{\rm c}_j, p^{\rm l}_j$ denote the center and the length of the reference point, respectively. 
2) For each decoder layer, we fix the sampling keys in deformable attention as $K=4$ evenly spaced positions over a specified interval from $p^{\rm c}_j-0.5 p^{\rm l}_j$ to $p^{\rm c}_j+0.5 p^{\rm l}_j$ to stabilize the network training. 
3) Without gIOU cost in bipartite matching, it is hard to accurately assign the target captions to event queries. We design the caption cost to mitigate this problem. Given any ground-truth caption $S_{j'} = \{w_{j't}\}_{t=1}^{M_{j'}}$ and any event query features $\tilde{q}_j$, we obtain the output probabilities $\{c^{\rm cap}_{jj't}\}_{t=1}^{M_{j'}}$ predicted by the captioning head with teacher forcing, where $M_{j'}$ denotes the caption length. The caption cost matrix is calculated by: $$(C_{\rm \rm cap})_{jj'}=\frac{1}{M_{j'}^{\gamma}} \sum_{t=1}^{M_{j'}} \log(c^{\rm cap}_{jj't}),$$
where $\gamma=2$ is the modulation factor of the caption length. The final cost matrix for bipartite matching is: 
\begin{equation}
  {C} = {C}_{\rm cap} + \alpha_{\rm cls} {L}_{\rm cls},
\end{equation}
where $\alpha_{cls}=0.5$ is the balance factor.

Based on the above modification, we train PDVC\_light with merely captioning loss on YouCook2. We choose the lightweight captioning head to ease the optimization difficulty. During inference, we directly use the reference points in the last layer as the predicted proposals.

\begin{figure}
    \begin{subfigure}{.235\textwidth}
        \includegraphics[width=\textwidth]{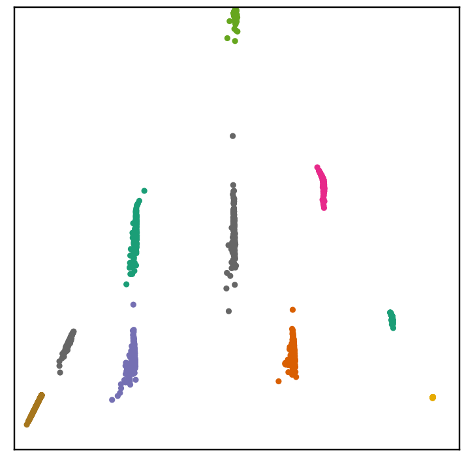}
        \caption{\scriptsize Predicted Proposals on ActivityNet Captions}
    \end{subfigure}
    \begin{subfigure}{.235\textwidth}
        \includegraphics[width=\textwidth]{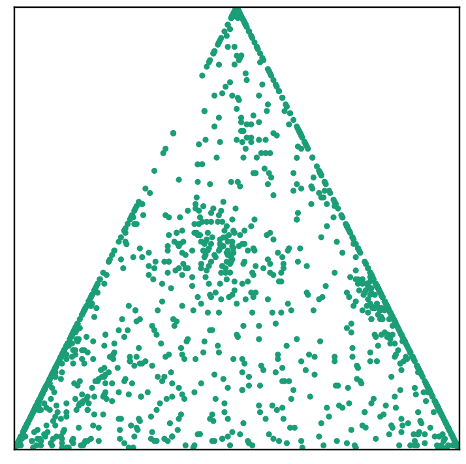}
        \caption{\scriptsize GT Proposals on ActivityNet Captions}
    \end{subfigure}
    
    \begin{subfigure}{.235\textwidth}
        \includegraphics[width=\textwidth]{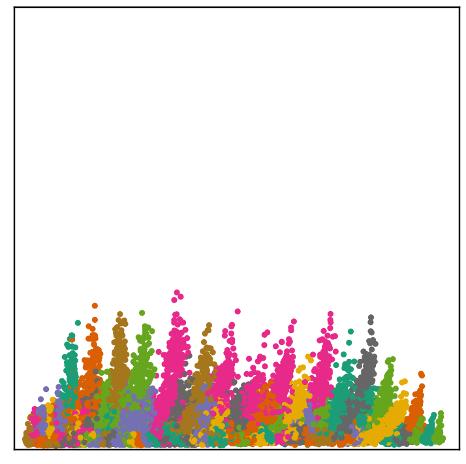}
        \caption{\scriptsize Predicted Proposals on YouCook2}
    \end{subfigure}
    \begin{subfigure}{.235\textwidth}
        \includegraphics[width=\textwidth]{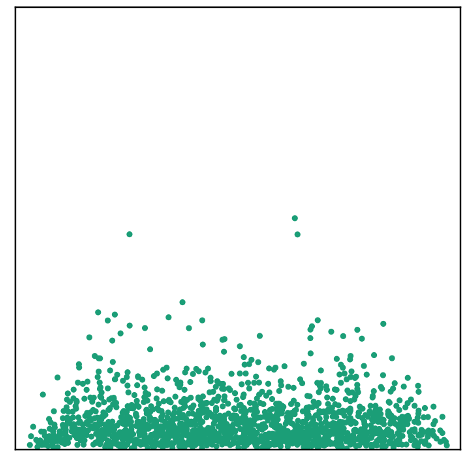}
        \caption{\scriptsize GT Proposals on YouCook2}
    \end{subfigure}
    \caption{The distribution of predicted proposals and ground-truth proposals. Horizontal and vertical axes represent the normalized center position and normalized length of proposals, respectively. For each dataset, we report the results of 200 randomly sampled videos on the validation set. The sub-figure~(a)/(c) contain 10/100 clusters with different colors, where each cluster corresponds to one event query.}
    \label{fig:prop} 
\end{figure}

\begin{figure*}
    \centering
    \includegraphics[width=1.0\textwidth]{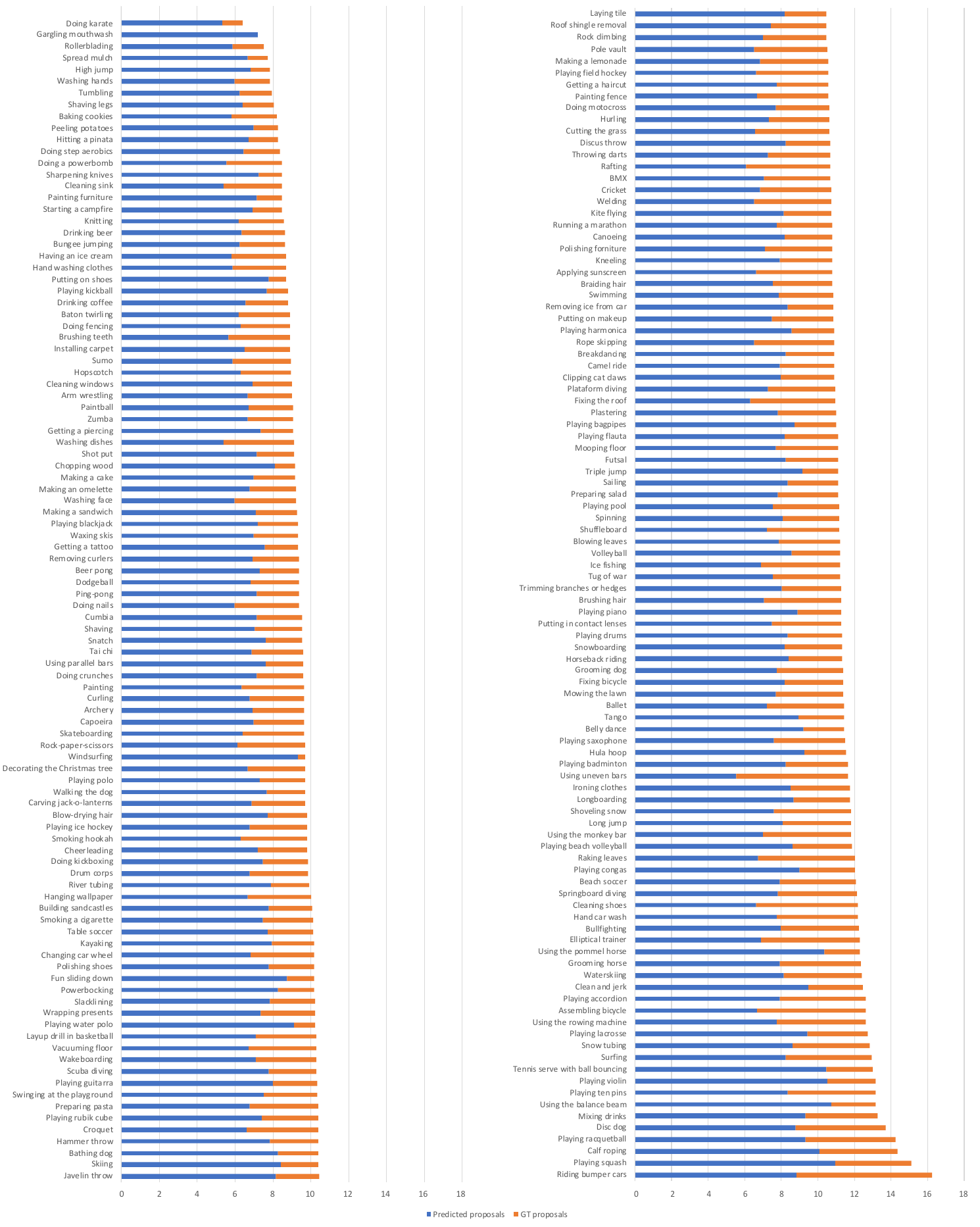}
    \caption{Dense captioning performance of PDVC on different activity classes. Activity labels are from the ActivityNet1.3 dataset~\cite{heil2015act}.}
    \label{fig:class}
\end{figure*}

\begin{figure*}
    \centering
    \includegraphics[width=1.0\textwidth]{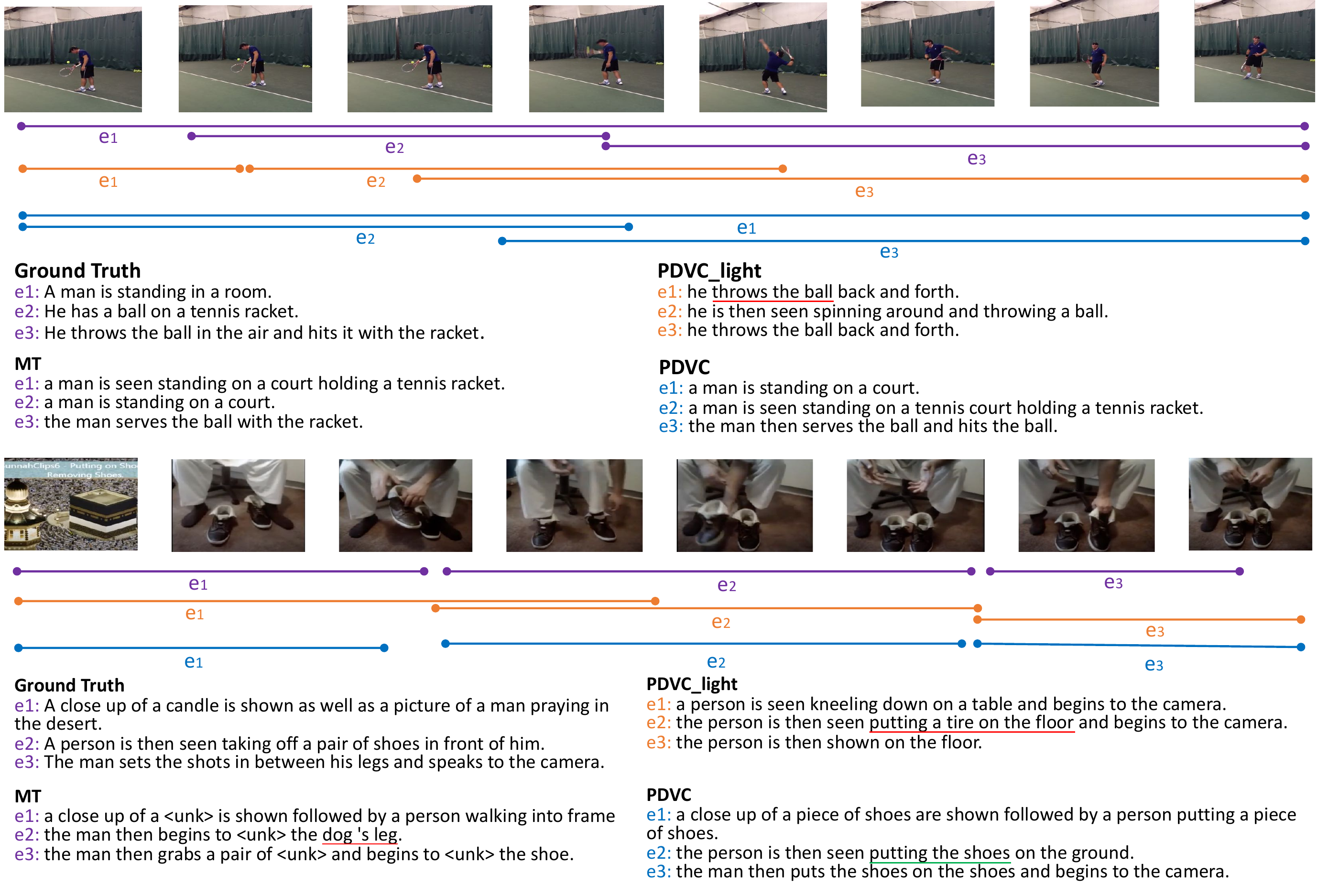}
    \caption{Visualization of predicted dense captions. Incorrect phases are underlined in red and the correct ones in green.}
    \label{fig:vis}
\end{figure*}

\subsection{Visualization}

\vspace{0.5em}
\noindent{\textbf{Predicted proposals.}}
We visualize the distribution of generated proposals of PDVC in Fig.~\ref{fig:prop}. For the ActivityNet Captions dataset, ground-truth proposals are distributed evenly across different positions and different lengths. However, for YouCook2, the length of most ground-truth proposals is relatively small (less than 25\% of the video duration). From the figure, we conclude that: 1) Each query describes a speciﬁc mode of the proposals’ location. 2) All queries can predict video-wide proposals with coherence and low redundancy and generate a similar distribution with ground truth. 3) Event queries serve as a location prior for localization tasks, which are trained to learn location patterns of events from human annotations. 

\vspace{0.5em}
\noindent{\textbf{Activity types.}}
The dense captioning performance of PDVC varies in different activity types. Fig.~\ref{fig:class} shows the METEOR score of PDVC with predicted/ground-truth proposals on 200 activity classes. Our model seems to generate more accurate captions with activities containing distinct scene cues or large objects, like ``riding bumper cars", ``playing squash", and ``calf roping". However, activities that rely more on fine-grained action cues or small objects tend to get a worse METEOR, like ``doing karate", ``gargling mouthwash", and ``rollerblading". It is promising to achieve a performance improvement to incorporate the fine-grained object features and a more powerful action recognition model.

\vspace{0.5em}
\noindent{\textbf{Temporally-localized captions.}}
Fig.~\ref{fig:vis} shows the generated captions with their temporal locations of different models. The captions of MT~\cite{zhou2018end} are generated based on ground-truth proposals, while PDVC\_light and PDVC are with predicted proposals. For the second video, MT and PDVC\_light misrecognize the shoes as a dog and a tire, respectively. Instead, PDVC can generate accurate and meaningful captions with predicted proposals, which verifies the effectiveness of the proposed parallel decoding mechanism and the captioning head with deformable soft attention.

\end{document}